%% file: root.tex
\documentclass[letterpaper, 10 pt, conference]{ieeeconf}  % Comment this line out if you need a4paper

\IEEEoverridecommandlockouts                              % This command is only needed if 
\usepackage{graphics} % for pdf, bitmapped graphics files
\usepackage{epsfig} % for postscript graphics files
\usepackage{mathptmx} % assumes new font selection scheme installed
\usepackage{times} % assumes new font selection scheme installed
\usepackage{amsmath,bm}
\usepackage{amssymb}  % assumes amsmath package installed
\usepackage{pifont}% http://ctan.org/pkg/pifont

\usepackage{import}
\usepackage{calrsfs}
\usepackage{mathtools,xparse}
\let\labelindent\relax
\usepackage[inline]{enumitem}
\usepackage[hidelinks]{hyperref}
\usepackage{booktabs}
\usepackage{transparent}

\DeclareMathAlphabet{\pazocal}{OMS}{zplm}{m}{n}
\DeclarePairedDelimiter{\abs}{\lvert}{\rvert}
\DeclarePairedDelimiter{\norm}{\lVert}{\rVert}
\newcommand{\loss}{\pazocal{L}}

\usepackage{color, xcolor}
\usepackage{cleveref}      % For /cref command
\usepackage{caption}
\usepackage[bottom]{footmisc}

\usepackage{array}
\newcolumntype{P}[1]{>{\centering\arraybackslash}p{#1}}

\newcommand{\etal}{\textit{et al.}~}

\newcommand{\cmark}{\ding{51}}%
\newcommand{\xmark}{\ding{55}}%
\newcommand{\conc}{\mathbin{\|}}

\newcommand{\method}{{EagerNet}} %if required, name of method

\graphicspath{{figures/}}

\usepackage[toc,acronym,nonumberlist]{glossaries}
\makeglossaries
\title{\LARGE \bf
6D Object Pose Estimation from Approximate 3D Models \\ for Orbital Robotics
}

\author{Maximilian Ulmer$^{1}$, Maximilian Durner$^{1,2}$, Martin Sundermeyer$^{1,2}$, Manuel Stoiber$^{1,2}$, and Rudolph Triebel$^{1,2}$% <-this % stops a space
\thanks{$^{1}$ Institute of Robotics and Mechatronics, German Aerospace Center (DLR), 82234 Wessling, Germany {\tt\small <first>.<second>@dlr.de}}%
\thanks{$^{2}$Department of Computer Science, Technical University of Munich (TUM), 85748 Garching, Germany}%
}

\begin{document}
\maketitle
\thispagestyle{empty}
\pagestyle{empty}

\input{abbreviations}
\begin{abstract}
 We present a novel technique to estimate the 6D pose of objects from single images where the 3D geometry of the object is only given approximately and not as a precise 3D model. To achieve this, we employ a dense 2D-to-3D correspondence predictor that regresses 3D model coordinates for every pixel. In addition to the 3D coordinates, our model also estimates the pixel-wise coordinate error to discard correspondences that are likely wrong. This allows us to generate multiple 6D pose hypotheses of the object, which we then refine iteratively using a highly efficient region-based approach. We also introduce a novel pixel-wise posterior formulation by which we can estimate the probability for each hypothesis and select the most likely one. As we show in experiments, our approach is capable of dealing with extreme visual conditions including overexposure, high contrast, or low signal-to-noise ratio. This makes it a powerful technique for the particularly challenging task of estimating the pose of tumbling satellites for in-orbit robotic applications. Our method achieves state-of-the-art performance on the SPEED+ dataset and has won the SPEC2021 post-mortem competition. 
 Code, trained models, and the used satellite model will be made publicly available.
\end{abstract}

%%%%%%%%%%%%%%%%%%%%%%%%%%%%%%%%%%%%%%%%%%%%%%%%%%%%%%%%%%%%%%%%%%%%%%%%%%%%%%%%
\input{sections/introduction}
\input{sections/related_work}
\input{sections/method}
\input{sections/experiments}

\input{sections/conclusions}

%%%%%%%%%%%%%%%%%%%%%%%%%%%%%%%%%%%%%%%%%%%%%%%%%%%%%%%%%%%%%%%%%%%%%%%%%%%%%%%%
%\section*{APPENDIX}

%\input{sections/acknowledgment}

%%%%%%%%%%%%%%%%%%%%%%%%%%%%%%%%%%%%%%%%%%%%%%%%%%%%%%%%%%%%%%%%%%%%%%%%%%%%%%%%

%References are important to the reader; therefore, each citation must be complete and correct. If at all possible, references should be commonly available publications.

\bibliographystyle{IEEEtran}
\bibliography{IEEEabrv,references}

\end{document}

%% file: abbreviations.tex
%A
%B
\newacronym{BCE}{BCE}{Binary Cross Entropy}
%C
\newacronym{CE}{CE}{Cross Entropy}
\newacronym{CNN}{CNN}{convolutional neural network}

%D
\newacronym{DL}{DL}{Deep Learning}
%E
\newacronym{ESA}{ESA}{European Space Agency}

%F
%G
\newacronym{GPU}{GPU}{graphics processing unit}
\newacronym{GPUs}{GPUs}{graphics processing units}
\newacronym{GAN}{GAN}{generative adverserial network}

%H
\newacronym{HIL}{HIL}{hardware-in-the-loop}
%I
%J
%K
%L
%M
%N
%O
%P
\newacronym{PnP}{PnP}{Perspective-n-Point}
%Q
%R
\newacronym{RoI}{RoI}{Region of Interest}
\newacronym{RANSAC}{RANSAC}{Random Sample Consensus}
%S
\newacronym{SPEC}{SPEC}{Satellite Pose Estimation Challenge}
%T
%U
%V
\newacronym{VOC}{VOC}{PASCAL Visual Object Classes}

%% file: sections/introduction.tex
\section{Introduction}
The field of orbital robotics is gaining traction, as humans' presence in space expands and the sustainability of the orbital environment as well as its infrastructure are becoming increasingly important. 
Especially for highly complex operations such as on-orbit servicing, active debris removal, and on-orbit assembly, accurate and high-performance determination of the target's pose is vital. 
Due to their lower mass, power consumption, and system complexity, pose estimation systems based solely on monocular cameras have recently emerged as an appealing alternative to systems based on active sensors. 
However, the orbit is an extremely challenging domain for camera-based algorithms due to hostile illumination, aggressive reflections, high contrast, and background noise. 
These conditions drastically deteriorate the performance of classical, shape-based algorithms due to obscured geometric features. %because they can obscure geometric features of the target.

Recently, pose estimation methods based on \gls{DL} have shown great capabilities %to deal with such visual effects 
in terrestrial scenarios~\cite{sundermeyer2023bop}.
However, the space domain remains hesitant to widely adopt such methods, due to high computational cost, availability of space-qualified \gls{GPUs}, software verification, and validation concerns. 
Moreover, learning-based methods often require large amounts of annotated training data.% to achieve good performance.
Therefore, the pose estimation community is drifting towards the use of synthetic training data which requires drastically less human effort~\cite{denninger_blenderproc_nodate,muller_photorealistic_2021}.

\begin{figure}
    \centering
    %\def\svgwidth{\columnwidth}
    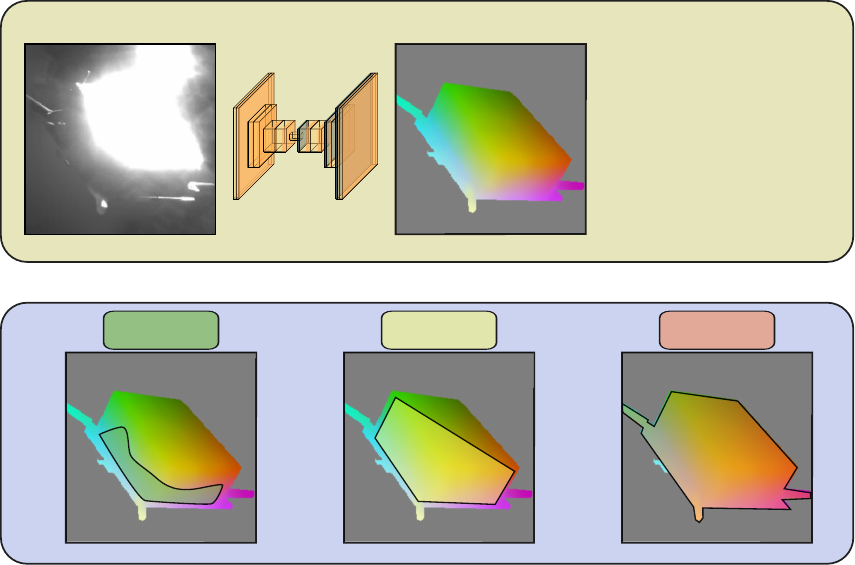
    \caption{We predict 2D-3D correspondences and a pixel-wise coordinate estimation error using a single \gls{CNN}. Then, we leverage the error prediction with different thresholds $\epsilon \in [\epsilon_1,\dots,\epsilon_n]$ to formulate multiple pose hypotheses using \gls{PnP}.}
    \label{fig:introduction}
\vspace{-3mm}
\end{figure}
To this end, the \gls{ESA} and Stanford University published the SPEED+ dataset~\cite{Park2021-jr} -- and hosted an accompanying competition -- designed to investigate the performance gap between training on synthetic and testing on real data in the space domain. 
The competition has seen avid participation from the community and a host of solutions have been submitted. 
\begin{figure*}
    \centering
    \def\svgwidth{\textwidth}
    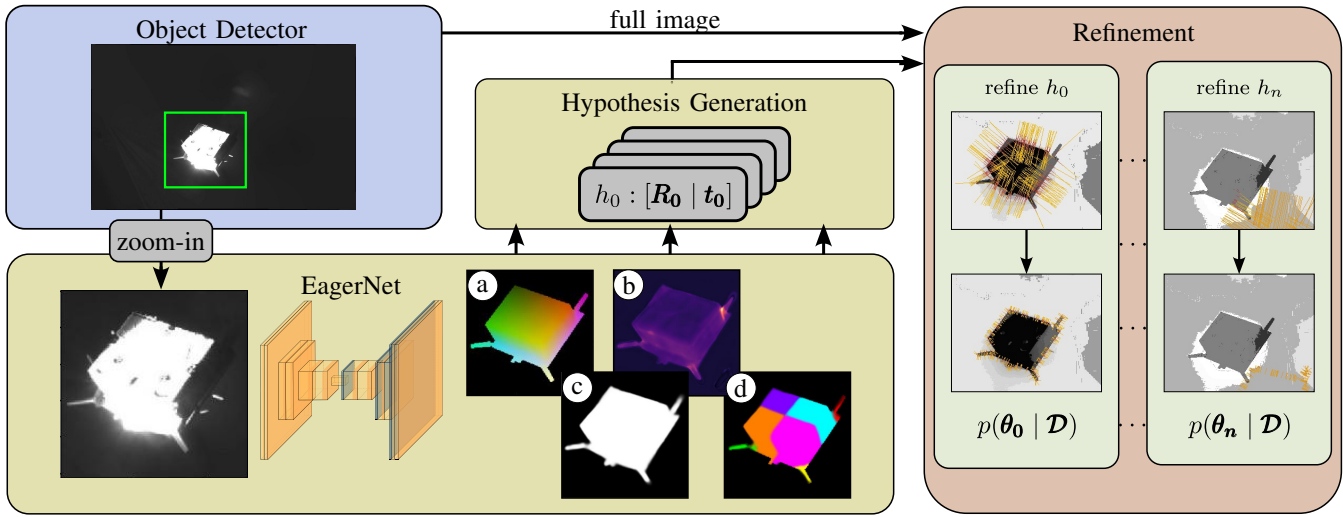
    \caption{Overview of our proposed framework using \method. Given an RGB image, an object detector estimates a \gls{RoI} which is input to \method. Then, it estimates several pixel-wise features: (a) normalized 3D model coordinates, (b) coordinate error (c) foreground confidence, and (d) surface region class labels. The predictions (a), (b), and (c) are used to formulate pose hypotheses using \gls{PnP}. At last, each hypothesis is optimized and a final probability per pose is calculated.}
    \label{fig:overview}
\vspace{-3mm}
\end{figure*}
Many of the best-performing methods~\cite{Perez-Villar_undated-ly, Wang2022-rs} rely on regressing sparse 2D-3D correspondences, called keypoints, from the input image to solve the 6D pose through a variant of the \gls{PnP} algorithm. 
While these methods generally perform well, they have some downsides. 
Namely, they require manual selection of keypoints and can struggle whenever a large number of keypoints is occluded or obscured due to other visual effects. 
Moreover, as shown in recent terrestrial benchmarks~\cite{sundermeyer2023bop} they fall behind dense correspondence methods in terms of accuracy.
In contrast, methods that densely establish 2D-3D correspondence can struggle with inaccurate models and symmetries.

In this work, we propose to overcome these limitations by establishing dense 2D-3D correspondences while simultaneously training the model to predict errors in its correspondence regression. 
In essence, this allows the pipeline to programmatically discard unreliable correspondences which can be a result of 3D model inaccuracies, occlusions, or harsh visual effects. 
Given the predicted error map, we generate multiple pose hypotheses (see Fig.~\ref{fig:introduction}) which are further refined. 
At last, we select the most probable pose hypothesis, given color statistics and learned features.  
These additions, significantly boost the performance of our dense correspondence-based approach outperforming all prior satellite pose estimation methods and achieving state-of-the-art on the competitive SPEED+ benchmark. 
To summarize, the contributions of this paper are
\begin{enumerate*}[label=\textbf{\arabic*)}]
    \item  the \underline{E}rror-\underline{a}ware, \underline{g}eometrically-guid\underline{e}d, co\underline{r}respondence network (EagerNet), a 6D pose estimation algorithm that is robust to 3D modelling errors and harsh visual conditions,
     \item a hybrid learning- and region-based 6D pose refinement, 
     \item error weighed multi-hypothesis generation and probabilistic pose selection, 
     \item state-of-the-art results on SPEED+~\cite{Park2021-jr} with ablations on the method and data, and 
     \item a study on the effects of 3D model quality on the TUD-L~\cite{hodan_bop_2018} dataset.
\end{enumerate*}

%% file: images/introduction.pdf_tex
%% Creator: Inkscape 1.2.2 (1:1.2.2+202212051550+b0a8486541), www.inkscape.org
%% Accompanies image file 'introduction.pdf' (pdf, eps, ps)
\begingroup%
  \makeatletter%
  \providecommand\color[2][]{%
    \errmessage{(Inkscape) Color is used for the text in Inkscape, but the package 'color.sty' is not loaded}%
    \renewcommand\color[2][]{}%
  }%
  \providecommand\transparent[1]{%
    \errmessage{(Inkscape) Transparency is used (non-zero) for the text in Inkscape, but the package 'transparent.sty' is not loaded}%
    \renewcommand\transparent[1]{}%
  }%
  \providecommand\rotatebox[2]{#2}%
  \newcommand*\fsize{\dimexpr\f@size pt\relax}%
  \newcommand*\lineheight[1]{\fontsize{\fsize}{#1\fsize}\selectfont}%
  \ifx\svgwidth\undefined%
    \setlength{\unitlength}{245.99998474bp}%
    \ifx\svgscale\undefined%
      \relax%
    \else%
      \setlength{\unitlength}{\unitlength * \real{\svgscale}}%
    \fi%
  \else%
    \setlength{\unitlength}{\svgwidth}%
  \fi%
  \global\let\svgwidth\undefined%
  \global\let\svgscale\undefined%
  \makeatother%
  \begin{picture}(1,0.66105067)%
    \lineheight{1}%
    \setlength\tabcolsep{0pt}%
    \put(0,0){\includegraphics[width=\unitlength,page=1]{introduction.pdf}}%
    \put(0.14069869,0.61801569){\color[rgb]{0,0,0}\transparent{0.87142199}\makebox(0,0)[t]{\lineheight{1.25}\smash{\begin{tabular}[t]{c}Input\end{tabular}}}}%
    \put(0.57454966,0.6156469){\color[rgb]{0,0,0}\transparent{0.87142199}\makebox(0,0)[t]{\lineheight{1.25}\smash{\begin{tabular}[t]{c}Coordinates\end{tabular}}}}%
    \put(0,0){\includegraphics[width=\unitlength,page=2]{introduction.pdf}}%
    \put(0.84275484,0.61563188){\color[rgb]{0,0,0}\transparent{0.87142199}\makebox(0,0)[t]{\lineheight{1.25}\smash{\begin{tabular}[t]{c}Error Prediction\end{tabular}}}}%
    \put(0.49780949,0.3207994){\color[rgb]{0,0,0}\transparent{0.87142199}\makebox(0,0)[t]{\lineheight{1.25}\smash{\begin{tabular}[t]{c}Hypothesis Generation\end{tabular}}}}%
    \put(0,0){\includegraphics[width=\unitlength,page=3]{introduction.pdf}}%
  \end{picture}%
\endgroup%

%% file: images/method.pdf_tex
%% Creator: Inkscape 1.2.2 (1:1.2.2+202212051550+b0a8486541), www.inkscape.org
%% Accompanies image file 'method.pdf' (pdf, eps, ps)
\begingroup%
  \makeatletter%
  \providecommand\color[2][]{%
    \errmessage{(Inkscape) Color is used for the text in Inkscape, but the package 'color.sty' is not loaded}%
    \renewcommand\color[2][]{}%
  }%
  \providecommand\transparent[1]{%
    \errmessage{(Inkscape) Transparency is used (non-zero) for the text in Inkscape, but the package 'transparent.sty' is not loaded}%
    \renewcommand\transparent[1]{}%
  }%
  \providecommand\rotatebox[2]{#2}%
  \newcommand*\fsize{\dimexpr\f@size pt\relax}%
  \newcommand*\lineheight[1]{\fontsize{\fsize}{#1\fsize}\selectfont}%
  \ifx\svgwidth\undefined%
    \setlength{\unitlength}{505.362854bp}%
    \ifx\svgscale\undefined%
      \relax%
    \else%
      \setlength{\unitlength}{\unitlength * \real{\svgscale}}%
    \fi%
  \else%
    \setlength{\unitlength}{\svgwidth}%
  \fi%
  \global\let\svgwidth\undefined%
  \global\let\svgscale\undefined%
  \makeatother%
  \begin{picture}(1,0.381483)%
    \lineheight{1}%
    \setlength\tabcolsep{0pt}%
    \put(0,0){\includegraphics[width=\unitlength,page=1]{method.pdf}}%
    \put(0.50777432,0.30220954){\color[rgb]{0,0,0}\transparent{0.99917799}\makebox(0,0)[t]{\lineheight{1.25}\smash{\begin{tabular}[t]{c}Hypothesis Generation\end{tabular}}}}%
    \put(0.49345524,0.36375254){\color[rgb]{0,0,0}\transparent{0.99917799}\makebox(0,0)[t]{\lineheight{1.25}\smash{\begin{tabular}[t]{c}full image\\\end{tabular}}}}%
    \put(0.16165294,0.35671093){\color[rgb]{0,0,0}\transparent{0.99917799}\makebox(0,0)[t]{\lineheight{1.25}\smash{\begin{tabular}[t]{c}Object Detector\end{tabular}}}}%
    \put(0.8437558,0.35471573){\color[rgb]{0,0,0}\transparent{0.99917799}\makebox(0,0)[t]{\lineheight{1.25}\smash{\begin{tabular}[t]{c}Refinement\end{tabular}}}}%
    \put(0.25858659,0.16482498){\color[rgb]{0,0,0}\transparent{0.99917799}\makebox(0,0)[t]{\lineheight{1.25}\smash{\begin{tabular}[t]{c}EagerNet\end{tabular}}}}%
    \put(0,0){\includegraphics[width=\unitlength,page=2]{method.pdf}}%
    \put(0.54988501,0.08798873){\color[rgb]{0,0,0}\makebox(0,0)[t]{\lineheight{1.25}\smash{\begin{tabular}[t]{c}d\end{tabular}}}}%
    \put(0,0){\includegraphics[width=\unitlength,page=3]{method.pdf}}%
    \put(0.35675498,0.16681436){\color[rgb]{0,0,0}\makebox(0,0)[t]{\lineheight{1.25}\smash{\begin{tabular}[t]{c}a\end{tabular}}}}%
    \put(0,0){\includegraphics[width=\unitlength,page=4]{method.pdf}}%
    \put(0.46642902,0.16519229){\color[rgb]{0,0,0}\makebox(0,0)[t]{\lineheight{1.25}\smash{\begin{tabular}[t]{c}b\end{tabular}}}}%
    \put(0,0){\includegraphics[width=\unitlength,page=5]{method.pdf}}%
    \put(0.42779431,0.09050843){\color[rgb]{0,0,0}\makebox(0,0)[t]{\lineheight{1.25}\smash{\begin{tabular}[t]{c}c\end{tabular}}}}%
    \put(0,0){\includegraphics[width=\unitlength,page=6]{method.pdf}}%
    \put(0.11631747,0.19811491){\color[rgb]{0,0,0}\transparent{0.99917799}\makebox(0,0)[t]{\lineheight{1.25}\smash{\begin{tabular}[t]{c}zoom-in\end{tabular}}}}%
    \put(0,0){\includegraphics[width=\unitlength,page=7]{method.pdf}}%
  \end{picture}%
\endgroup%

%% file: sections/related_work.tex
\section{Related Work}
\subsection{Monocular Satellite Pose Estimation} 
Recently, the space community started adopting \gls{DL}-based pose estimation methods. 
In~\cite{sharma2018pose}, one of the earliest adoptions, a pre-trained backbone is used to directly classify discretized poses. 
Around the same time, the SPEED dataset~\cite{Kisantal2019-yp} was published and the \gls{SPEC} was organized for the first time.
It saw a diverse set of learning-based approaches, such as probabilistic rotation estimation with a soft classifier~\cite{Proenca2020-jg}. 
Notably, many competitors used keypoint-based methods~\cite{Park2019-fp, Chen2019-iw} which have been widely used in the domain since then~\cite{black2021real, pasqualetto2020cnn}.
Shortly after, \cite{Hu2021-pl} published the SwissCube dataset which is a purely synthetic satellite pose estimation dataset, created with a physics-based renderer. 
The dataset focuses on the wide-depth-range scenario and the authors propose a single-stage hierarchical keypoint-based method that can deal with large-scale variations.
In 2021, the \gls{SPEC} competition was organized again, based on the SPEED+~\cite{Park2021-jr} dataset which introduced a vastly larger training set, empowering new \gls{DL} approaches. 
Yet, many of the top competitors still employed keypoint-based approaches. 
For instance, \cite{Perez-Villar_undated-ly} estimates keypoint location and depth in two decoder heads, and \cite{Wang2022-rs} augments their training images with a \gls{GAN}~\cite{goodfellow2020generative} to bridge the domain gap.

In contrast to these sparse correspondence methods, we predict 2D-3D correspondences densely per pixel and, to the best of our knowledge, present the first such method for satellite pose estimation.
In \cite{Li2019-tq}, dense correspondences are applied to estimate the rotation of objects and utilize a second head to predict the object translation, while \cite{Park2019-ny} predicts correspondence errors and uses a \gls{GAN} to enhance target predictions, and ~\cite{Wang2021-sj} uses a neural network-based \gls{PnP} to estimate the pose from the predicted dense correspondences. 
However, in contrast to these methods, we only assume to have access to an approximate 3D CAD model for training and explicitly handle inaccuracies in the 3D model.

\subsection{6D Pose Refinement}
After estimating a global pose, a local search can be used to further invest computational resources and improve estimation accuracy. DeepIM~\cite{Li2018-fy} uses \gls{DL} to refine a pose hypothesis by iteratively estimating a relative pose between a rendered object view in the current pose and the observed object view. 
CosyPose~\cite{Labbe2020-mr} improves on this idea with an optimized architecture and rotation parameterization. 
Similarly, \cite{Lipson2022-pq} uses the "render-and-compare" approach for refinement but instead of directly regressing the pose they apply bidirectional depth-augmented \gls{PnP} to estimate a pose update. 

This work uses a classical, region-based method to refine poses based on~\cite{Stoiber2020-gb, Stoiber2022-ka}. 
This class of methods differentiates between a foreground and background area using image statistics. 
Then, an optimizer searches for the model pose that best explains this segmentation. 
In~\ref{subsec:multi_hypothesis} we show that global color statistics alone are not enough to refine object poses in the harsh visual conditions of space. 
We, therefore, introduce a novel formulation to apply \gls{DL}-based features to the refinement step that can cope with the most aggressive of lighting conditions.

%% file: sections/method.tex
\section{Method} 
Given a 3D model of a rigid object, our goal is to estimate its 6D pose $[\bm{R}|\bm{t}] \in SE(3)$, where $\bm{R}$ and $\bm{t}$ represent the rotation matrix and translation vector respectively.
First, we employ a 2D object detector that estimates a \gls{RoI} in an input image. 
Then, we propose an asymmetric encoder-decoder architecture\footnote{The encoder part is relatively larger than the decoder part.}, called \method, that predicts 2D-3D correspondences between pixel space and object coordinates. 
The correspondences from the \gls{RoI} are transformed to the original image to directly estimate the relative camera-to-object transformation using the well-established \gls{PnP} algorithm~\cite{Lepetit_undated-le}.
Inspired by~\cite{Park2019-ny}, our model additionally predicts the error of estimated model coordinates to exclude erroneous correspondences in the \gls{PnP} step.
This enables increased robustness in cases where the CAD models are imperfect due to 3D reconstruction, modeling errors, or other deteriorating circumstances. 
In addition, we propose the addition of surface region segmentation as an auxiliary task to address the issue of object symmetries~\cite{Hodan2020-dz}, similar to~\cite{Wang2021-sj}. 
To further improve the estimated pose, we use the predicted error map to formulate different object pose hypotheses.
At last, we refine all hypotheses using a probabilistic refinement~\cite{Stoiber2020-gb} and select the pose estimate with the highest probability. 

\subsection{Learning-based Monocular 6D Pose Estimation}
\label{subsec:pose_estimation}
As visualized in Figure~\ref{fig:overview} we decouple detection from pose estimation and input a cropped, square RGB image $\bm{I} \in \mathbb{N}^{h\text{x}w\text{x}3}$ of a detected object to our network.
In the following, we elaborate further on each particular output.

\subsubsection{Dense Correspondence Regression}\label{meth:dense}
Our goal is to regress the coordinates of model points $\bm{q}=[x,y,z]^\mathsf{T}\in \mathbb{R}^3$ on the surface of an object in its inertial frame for each object image coordinate $\bm{p}= [u, v]^\mathsf{T}\in\mathbb{N}^2$. 

Hence, \method~outputs a matrix $\widehat{\bm{I}}_{\tilde{q}} \in \mathbb{R}^{h\text{x}w\text{x}3}$, where each pixel location corresponds to the normalized object's coordinates $\bm{\tilde{q}}$.
Each channel represents one of the three coordinate axes.
To generate the coordinate target $\bm{I}_{\tilde{q}}$ we render the object using its approximate 3D model to obtain $\bm{I_{q}}$ and normalize it. 
Simultaneously, we use the rendering to determine the target object mask $\bm{I}_{o} \in \mathbb{Z}_{2}^{h\text{x}w}$. 
It should be noted, that these renderings are based on the approximate CAD model available and do not have to be the ground truth.

We apply the $L_1$ loss only on the pixel within the object mask ${\bm{I}_o}$:
\begin{equation}
	\loss_{\tilde{q}} = \norm{\bm{I}_{o} \odot (\widehat{\bm{I}}_{\tilde{q}} - \bm{I}_{\tilde{q}})}_{1} ,
\end{equation}
where $\odot$ denotes element-wise multiplication. 

In addition, our method also outputs a confidence map $\widehat{\bm{I}}_o \in \mathbb{R}^{h\text{x}w}$ where each pixel represents a score whether it belongs to the object or not.

To learn the object segmentation, we employ the \gls{BCE} loss
\begin{equation}
	\loss_{o} = \gls{BCE}(\widehat{\bm{I}}_o, \bm{I_{o}}) .
\end{equation}

\subsubsection{Coordinate Prediction Error Estimation}
Given both maps, one could directly estimate an object's 6D pose by applying a threshold on the confidence $\widehat{\bm{I}}_o$ and forwarding the corresponding 3D points of $\widehat{\bm{I}}_{\tilde{q}}$ with their corresponding 2D locations to the \gls{PnP} algorithm.

In this case, the PnP post-processing exclusively relies on the predicted object mask and equally considers all estimated corresponding model coordinates.
However, due to deteriorating visual effects, occlusions, or modeling errors many of those coordinates can be erroneous (see~\ref{subsubsec:exp:error}).
Although the \gls{RANSAC} voting scheme employed with most \gls{PnP} approaches handles outliers effectively, these prediction errors accumulate and deteriorate the final rotation and especially translation estimates. 
This is one of the reasons, why Li~\etal\cite{Li2019-tq} propose an additional network to directly predict the translation.
In this work, we address this issue by excluding possibly erroneous correspondences, inspired by~\cite{Park2019-ny}.

We add an additional output channel which is trained to predict the $L_1$ error of the predicted model coordinates at each foreground pixel location $\bm{I}_e= \norm{\bm{I}_{o} \odot (\widehat{\bm{I}}_{\tilde{q}} - \bm{I}_{\tilde{q}})}_{1}$ without reducing it.
We employ a bounded mean squared error as the training loss 
\begin{equation}
	\loss_{e} = min(\norm{\widehat{\bm{I}}_e - \bm{I}_e}^2_2, 1)
\end{equation}
for the predicted error map $\widehat{\bm{I}}_e$.
As our experiments show (cf.~\ref{subsubsec:exp:error}), this extra output channel does not reduce the coordinate prediction performance, even when learning with high-quality CAD models, and without requiring an extra network head for translation estimation.

Further, since $\bm{I}_e$ is calculated online and only requires $\bm{I}_o$ and $\bm{I}_{\tilde{q}}$, no additional label generation or data-loading logic is required such that any existing 2D-3D dense correspondence method can be extended with only a few lines of code.
% We will show in \todo{EXPERIMENTS} that adding this extra target to training does not deteriorate the coordinate prediction performance, even when learning with excellent CAD models. 
% In addition, the error prediction module comes with very little computational and software development overhead. 
% It can be added to any 2D-3D dense correspondence method in a few lines of code. 
% Since the target is calculated online, no additional data-loading logic is required.

\subsubsection{Surface Region Segmentation}
Satellites and many other objects can exhibit strong symmetries which can drastically impair the performance of pose estimation methods~\cite{Sundermeyer2018-fl}. 
Correspondence-based methods map local features in the input to 3D model coordinates and hence assume a one-to-one relationship. 
In the case of many-to-many correspondences, these methods trade-off possible locations and can return the average, which is often non-viable~\cite{Hodan2020-dz}. 

To this end, we add an auxiliary task that learns to classify sub-regions on the target object, inspired by~\cite{Wang2021-sj, Hodan2020-dz}.
We cast this as a multi-class segmentation and create $n$ region classes applying farthest-point sampling.

In essence, we calculate the $n$ model points farthest away from each other. 
Then, we classify each surface point depending on which farthest point it is closest to.
The resulting segmentation map is one-hot encoded which leads to the training target $\bm{I}_r \in \mathbb{Z}_{2}^{h\text{x}w\text{x}n}$, as described in~\cite{Hodan2020-dz}. % with one channel per surface region. 

Similar to $\loss_{\tilde{q}}$, the related loss $\loss_{r}$ only considers foreground pixels %similar to the error and coordinates loss functions 
and uses \gls{CE} 
\begin{equation}
	\loss_{r} = \gls{CE}(\bm{I}_o \odot \widehat{\bm{I}}_r, \bm{I}_r) .
\end{equation}

% The surface region segmentation $\widehat{\bm{I_r}}$ is only used during training and not used in the \gls{PnP} step.

\noindent\textbf{Training.} The total loss, consisting of coordinates $\loss_{\tilde{q}}$, foreground segmentation $\loss_{o}$, error prediction $\loss_{e}$, and surface region segmentation $\loss_{r}$ loss is 
\begin{equation}
	\loss = \alpha \cdot \loss_{\tilde{q}} + \beta \cdot \loss_{o} +  \gamma \cdot \loss_{e} + \delta \cdot \loss_{r} ,
\end{equation}

scaled by weights $\alpha, \beta, \gamma, \delta$. 
We evaluate different backbones and decoders and detail those in the experiment section. 

\subsection{Learned Region-based Pose Refinement}
In this work, we build on the probabilistic, region-based approach developed in \cite{Stoiber2020-gb, Stoiber2022-ka}.
In general, region-based methods use image statistics to differentiate between a foreground that corresponds to the object and a background.
They then try to find the pose that best explains this segmentation.
While region-based techniques work very well in a wide variety of scenarios, many methods only use simple color histograms to differentiate between the object and the background.
Thus, such approaches require distinct RGB colors to be viable.
To overcome this limitation and use region-based refinement in the hard visual conditions of orbital pose estimation, where color alone does not provide strong information, we incorporated the learned features from Section~\ref{meth:dense}.
In addition to learned color, they are able to include information from global object appearances such as texture or geometry.

During the optimization, information is only considered along so-called correspondence lines.
Correspondence lines are defined by a center $\bm{c}\in\mathbb{R}^2$ and a normal vector $\bm{n}\in\mathbb{R}^2$ that are projected from the view of a sparse viewpoint model closest to the current pose estimate.
Given a line coordinate $r\in \mathbb{R}$ and the corresponding image coordinate $\bm{p} = \bm{c}+r\bm{n}$, a correspondence line $\bm{l}$ maps a scalar $r\in \mathbb{R}$ to an image value $\bm{\tau} = \bm{l}(r)=\bm{I}(\bm{p})$, with $\bm{\tau} \in \{0,...,255\}^3$.
Like other approaches, we approximate the probability distributions $p(\bm{\tau} \mid m_\textrm{f})$ and $p(\bm{\tau} \mid m_\textrm{b})$ of foreground and background colors, with color histograms.
They are calculated by sampling pixels along correspondence lines on either side of the contour.
Given a specific pixel color $\bm{\tau}$, the distributions can be used to calculate the color-based pixel-wise posteriors of the foreground and background model $m_\textrm{f}$ and $m_\textrm{b}$ as
\begin{equation} \label{eq:ref:pwp}
	p(m_i \mid  \bm{\tau}) = \frac{p(\bm{\tau}\mid m_i)}{p(\bm{\tau} \mid  m_\textrm{f}) + p(\bm{\tau} \mid  m_\textrm{b})} , \quad i\in\{\textrm{f}, \textrm{b}\}.
\end{equation}

In general, color histograms alone can be brittle if the foreground and background exhibit similar color statistics.
We therefore incorporate the learned confidences $p_{\psi}(m_f \mid \bm{p})=\bm{\widehat{I}}_o(\bm{p})$ and $p_{\psi}(m_b \mid \bm{p})=1-\bm{\widehat{I}}_o(\bm{p})$, parameterized by the neural network weights $\psi$, from the pose estimator as salient features.
In order to retain information from both sources, we average the color-based pixel-wise posteriors and the learned confidences as follows
\begin{equation} \label{eq:method:pwp_theta}
	p_i = \frac{1}{2}(p(m_i \mid  \bm{\tau}) + p_{\psi}(m_i \mid \bm{p})), \quad i\in\{\textrm{f}, \textrm{b}\}.
\end{equation}
Due to the overconfidence of learned segmentation maps, most values in $\widehat{\bm{I}}_o$ are either $0$ or $1$. 
As a result, we cannot treat the learned confidence as a probability, otherwise, the learned confidence dominates the calculation of the probabilities.

Like in our previous work~\cite{Stoiber2022-lq}, we combine measurements from individual pixels along the correspondence line $\bm{l}$ to compute the probability of the contour location $d\in\mathbb{R}$ as follows
\begin{equation}
	p(d\mid \bm{\pazocal{D}}_{\bm{l}} ) \propto \prod_{r\in\omega}h_\textrm{f}(r-d)p_\textrm{f}(r) + h_\textrm{b}(r-d)p_\textrm{b}(r),
\end{equation}
with the correspondence line domain $\omega$  and the smoothed step functions $h_\textrm{f}$ and $h_\textrm{b}$.
Detailed information about those functions can be found in \cite{Stoiber2020-gb, Stoiber2022-ka}.
The contour location $d$ relative to the center $\bm{c}$ depends on the 6D pose variation $\bm{\theta}$ and can be calculated as follows
\begin{equation}
	d(\bm{\theta}) = \bm{n}^\top\big(\bm{\pi}(\bm{q}(\bm{\theta}))-\bm{c}\big),
\end{equation}
where $\bm{q}$ is the 3D model point from the sparse viewpoint model that was used to establish $\bm{c}$, given in the camera frame, and $\bm{p}=\bm{\pi} (\bm{q})$ denotes the pinhole camera model. 

Finally, information from valid correspondence lines is combined to estimate the probability of the pose variation
\begin{equation} \label{eq:method:probability}
	p(\bm{\theta}\mid \bm{\pazocal{D}}) \propto \prod_{i=1}^{n}p(d_{i}(\bm{\theta})\mid\bm{\pazocal{D}}_{{\bm{l}}i} )^\frac{s_\textrm{h} s^2}{{\sigma_\textrm{r}}^2 {\bar{n}_i}^2},
\end{equation}
with the scale $s\in \mathbb{N}^+$, the user-defined standard deviation $\sigma_\textrm{r}$, the slope parameter $s_\textrm{h} \in \mathbb{R}^+$, and the normal component $\bar{n}_i=\norm{\bm{n}_i}_{max}$.
While the standard deviation can be used to define the overall uncertainty, all other parameters are part of a scale-space formulation that improves efficiency.
For a detailed explanation, please refer to~\cite{Stoiber2020-gb, Stoiber2022-ka}.

To refine pose estimates, Eq.~(\ref{eq:method:probability}) is iteratively maximized using Newton optimization with Tikhonov regularization, considering both learned and color-based region information.
In addition, the formulation also allows the calculation of a probability that assesses the confidence of a particular estimate. 

\subsection{Multi-Hypothesis Testing}
\label{subsec:multi_hypothesis}
Given a probabilistic formulation that enables the comparison of different pose estimates in Eq.~(\ref{eq:method:probability}), we can now utilize the error prediction map $\widehat{\bm{I}}_e$ more efficiently. 

If the error threshold is just set to a static value, we have to find a trade-off between very accurate estimates and being robust to difficult samples. 
Consider the case, that the model is very confident about its current correspondence prediction. 
In that case, we only want to take the very best 2D-3D correspondences into account for \gls{PnP}. 
In contrast, if the visual conditions are challenging it is often beneficial to set the threshold to a higher value and consider more correspondences. % for the pose estimation. 

Using Eq.~\ref{eq:method:probability} of the refiner, we are able to estimate poses at different error thresholds $\varepsilon \in [0,1]$, refine each individual pose, and subsequently, compare the resulting probability. 
In the end, we choose the pose estimate with the highest probability, given the color statistics and learned confidence map $\widehat{\bm{I}}_o$.
\begin{figure*}[t]
    \centering
    \def\svgwidth{\textwidth}
    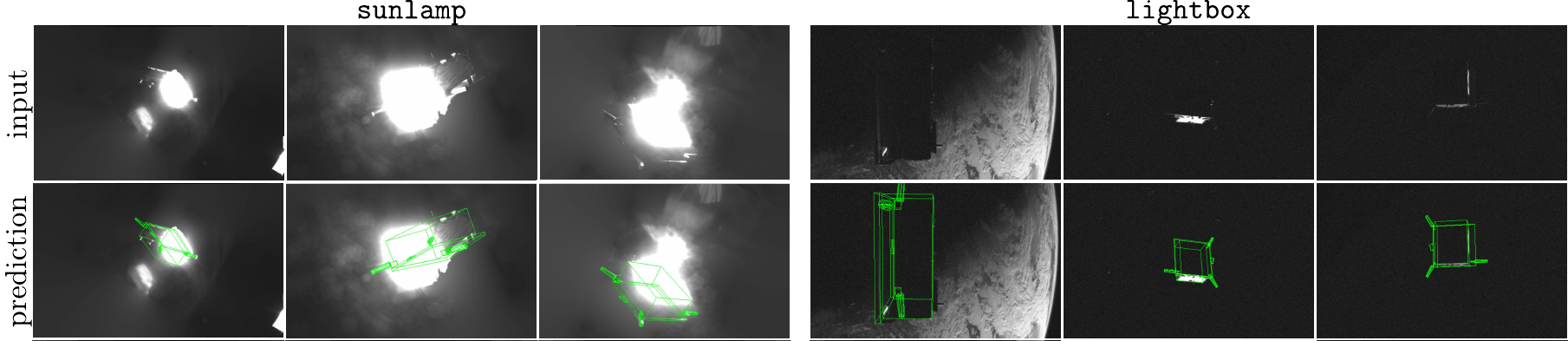
    \caption{Qualitative results of \method~on SPEED+~\cite{Park2021-jr} test samples. The top row depicts the input image while the bottom one shows a wireframe model with an estimated pose projected on the image. The left and right areas show samples from the \texttt{sunlamp} and \texttt{lightbox} categories respectively.}
    \label{fig:speedplus_samples}
\vspace{-4mm}
\end{figure*}

%% file: images/examples.pdf_tex
%% Creator: Inkscape 1.2.2 (1:1.2.2+202212051550+b0a8486541), www.inkscape.org
%% Accompanies image file 'examples.pdf' (pdf, eps, ps)
\begingroup%
  \makeatletter%
  \providecommand\color[2][]{%
    \errmessage{(Inkscape) Color is used for the text in Inkscape, but the package 'color.sty' is not loaded}%
    \renewcommand\color[2][]{}%
  }%
  \providecommand\transparent[1]{%
    \errmessage{(Inkscape) Transparency is used (non-zero) for the text in Inkscape, but the package 'transparent.sty' is not loaded}%
    \renewcommand\transparent[1]{}%
  }%
  \providecommand\rotatebox[2]{#2}%
  \newcommand*\fsize{\dimexpr\f@size pt\relax}%
  \newcommand*\lineheight[1]{\fontsize{\fsize}{#1\fsize}\selectfont}%
  \ifx\svgwidth\undefined%
    \setlength{\unitlength}{520.92797852bp}%
    \ifx\svgscale\undefined%
      \relax%
    \else%
      \setlength{\unitlength}{\unitlength * \real{\svgscale}}%
    \fi%
  \else%
    \setlength{\unitlength}{\svgwidth}%
  \fi%
  \global\let\svgwidth\undefined%
  \global\let\svgscale\undefined%
  \makeatother%
  \begin{picture}(1,0.21704422)%
    \lineheight{1}%
    \setlength\tabcolsep{0pt}%
    \put(0,0){\includegraphics[width=\unitlength,page=1]{examples.pdf}}%
  \end{picture}%
\endgroup%

%% file: sections/experiments.tex
\section{Experiments}
In this section, we first detail our experimental setup with which we competed in the post-mortem \gls{SPEC} competition and achieved state-of-the-art performance. In addition, we perform extensive experiments on the SPEED+~\cite{Park2021-jr} and TUD-L~\cite{hodan_bop_2018} datasets to investigate various aspects of our pose estimation system.

\subsection{SPEED+ Satellite Pose Estimation Challenge}
The Next-Generation \underline{S}pacecraft \underline{P}os\underline{E} \underline{E}stimation \underline{D}ataset (SPEED+) was the subject of the 6D pose estimation challenge %\footnote{https://kelvins.esa.int/pose-estimation-2021/home/} 
\gls{SPEC} organized by the \gls{ESA} and Stanford University. 
After its inception in 2021, the official part of the challenge concluded on 03/31/2022. Afterward, for post-mortem competition, the leaderboards remained open and competitors were able to submit results before the competition ultimately closed on 12/31/2022. 

The purpose of the competition was to determine the current state-of-the-art performance on the challenging task of satellite pose estimation as well as closing the Sim2Real gap. 
Hence, the training and validation samples of the dataset are $59960$ computer-rendered images of the Tango satellite annotated with the 6D pose.
The test set contains $9531$ \gls{HIL} mockup-satellite images split into two subsets %-- called \texttt{lightbox} and \texttt{sunlamp} -- 
with distinct visual conditions.
The \texttt{lightbox} split simulates albedo conditions using light boxes with diffuser plates and the \texttt{sunlamp} split entails a high-intensity sun simulator to mimic direct homogeneous light from the sun~\cite{Park2021-jr}.

\subsubsection{Synthetic Training Data}
\label{subsubsec:exp:data}
Our work is inspired by the issue of missing the satellite's CAD model but requiring one to train the proposed network architecture.
%The dataset does not include a ground truth CAD model of the satellite which motivates our approach to the problem.
To this end, we established a coarse wireframe model of the satellite by inspection and used this faulty model in combination with the 6D pose to generate the missing annotations (i.e. $\bm{I_{\tilde{q}}}, \bm{I}_o, \bm{I}_r$) for the original training images. 
In addition, we rectified each sample and reduced the resolution to $640 \times 400$ pixels.

\subsubsection{Closing the Sim2Real Gap} 
\label{subsubsec:exp:sim2real}
A central obstacle of the dataset is closing the Sim2Real gap. 
In this work, we focus on data augmentations which have long been a standard practice to train neural networks~\cite{devries2017improved} and are one of the simplest, yet effective steps to close the domain gap. 
%We augmented each training image by an array of randomized augmentations, the full stack of augmentations with all values can be found in the github repository. 
We applied an array of randomized augmentations to each training image\footnote{For details we would like to refer to our GitHub repository (soon available)}. %, the full stack of augmentations with all values can be found in the github repository.
Besides common augmentations (e.g. add, multiply, invert, brightness, blur, gaussian noise, and dropout)~\cite{Sundermeyer2018-fl} we use ~\gls{DL}-based style augmentations~\cite{Jackson2018-ye}. In addition, we crafted space-specific augmentations to tackle the unique effects seen in the test data. 
For instance, we implemented a synthetic specular reflection module that searches for connected regions in the image that are brighter than the rest of the image and adds a very bright blurry structure on top. This structure resembles features similar to specular reflection seen in orbit. 
Simultaneously, we make the rest of the image darker to mimic the exceedance of the dynamic range of the camera. 
This dramatically improves the prediction performances on the \texttt{sunlamp} test set (see ~\ref{exp:speedplus_ablation:domain_gap}). 

\subsubsection{Object Detection and Bounding Box Refinement} \label{subsubsec:exp:object_detection}
In the scope of the challenge, we trained an off-the-shelf object detector to predict 2D bounding boxes obtained by our created mask annotations $\bm{I}_o$. 
We used a Mask R-CNN~\cite{He2017-ko} with a pre-trained ResNet-50~\cite{He2015-hv} backbone and similar augmentations as described in~\ref{subsubsec:exp:sim2real}. 
The detector was trained for $130$ epochs using stochastic gradient descent with a learning rate of $0.01$, momentum $0.9$, and weight decay $0.0005$. In addition, we decayed the learning rate by $0.1$ every $3$ epochs. 
For two-stage approaches, the performance of 6D pose estimation algorithms correlates with the quality of the 2D object detection~\cite{sundermeyer2023bop}. % performance. 
Hence, we propose an iterative bounding box refinement during inference. % step for our approach. 
%For each sample, we first used the bounding box from the detector and predicted the error map $\bm{\widehat{I_e}}$. 
%If the standard deviation of the predicted error is below a threshold, we assume that the model is certain to a degree about its prediction. 
%In that case, we use the wireframe from ~\ref{subsubsec:exp:data} to change the bounding box such that it fits the model perfectly. 
We render the model wireframe (cf. \ref{subsubsec:exp:data}) if the standard deviation of the predicted error map is below a threshold and update the bounding box. 
% Thus, we exploit this value to determine good predictions.
% For samples below the threshold, we render the wireframe (cf. \ref{subsubsec:exp:data}) and update the bounding box.

\subsubsection{Test Set Augmentations}
%In addition to bounding box refinement, we added another step to trade-off extra compute for higher performance in the scope of the challenge. 
%We observed that pose estimation results were not identical when rotating input images, probably due to biases in the training data. 
Another empirical finding which leads to a \gls{SPEC} specific method adaptation was deviating results if the input images were rotated.
We hypothesize that this is caused by a bias in the training data.
Thus, we create an ensemble and forward each sample four times to our model, rotated by $\varrho \in [0, \frac{\pi}{2}, \pi$, $\frac{3\pi}{2}]$, respectively. 
After prediction, the rotations of all outputs are reversed. 

We generate $n_\varepsilon$ hypothesis (cf.~\ref{subsec:multi_hypothesis}) for each rotation, leading to a total of $4 \cdot n_\varepsilon$ hypotheses refinements
At last, we average the four resulting confidence maps $\widehat{\bm{I}}_o = \frac{1}{4}\sum_{i=1}^4 \widehat{\bm{I}}_{o,i}$ before refining each individual hypothesis.

\subsubsection{Training} \label{subsubsec:exp:training}
The network architecture is inspired by \cite{liu2022gdrnpp_bop}, a state-of-the-art 6D pose estimation method. As the backbone, we use ConvNeXt-Base~\cite{Liu_undated-mg}, pre-trained on ImageNet-1k. We extend the decoder of~\cite{liu2022gdrnpp_bop} by two additional upsampling steps but keep the kernel size of $3$ and feature dimensions of $256$ for each convolution the same. 
The group normalization~\cite{Wu2018-re} and GELU~\cite{Hendrycks2016-hm} activation are applied to every layer, except the output layer. In the last layer, we apply the sigmoid activation to $\widehat{\bm{I}}_o$ and $\widehat{\bm{I}_r}$. 

The model is trained for $2\mathrm{e}6$ steps with a batch size of $32$ using the Adam~\cite{Kingma2014-hu} optimizer with a base learning rate of $1\mathrm{e}{-3}$. The learning rate is warmed up for the first $1000$ steps and annealed after $72\%$ of the training is complete with a cosine scheduler~\cite{Wang2021-sj}.
We predict $8$ surface regions for $\bm{I}_r$ and scale the loss with weights $\alpha, \beta, \gamma = 1$ and $\delta=0.1$.

% We predict $8$ surface regions for $\bm{I}_r$. Therefore, the complete target tensor is composed of $\bigl[\bm{I_{\tilde{q}}} \conc \bm{I}_o \conc \bm{I}_e \conc \bm{I}_r\bigr] \in \mathbb{R}^{256 \times 256 \times 13}$ and scale the loss with weights $\alpha, \beta, \gamma = 1$ and $\delta=0.1$.

\subsubsection{Evaluation Metrics}
In the scope of the SPEED+ challenge, translational and rotational errors are used to evaluate a 6D pose~\cite{Park2021-jr}. 
The normalized position error $e_{\bm{t}}$ and rotation error $e_{\bm{R}}$ are defined as
\begin{equation}
    e_{\bm{t}} = \frac{\norm{\bm{\hat{t}} - \bm{t}}}{\bm{t}} , ~~ e_{\bm{R}} = 2\cdot \arccos{\abs{\langle \bm{\hat{a}}, \bm{a} \rangle}}
\end{equation}
% \begin{equation}
%     e_{\bm{R}} = 2\cdot \arccos{\abs{\langle \bm{\hat{a}}, \bm{a} \rangle}} ,
% \end{equation}
% \begin{equation}
%     e_{\text{pose}} = e_{\bm{t}} + e_{\bm{R}},
% \end{equation}
with quaternion representation $\bm{a}$ of rotation matrix $\bm{R}$. 
If the errors are below the calibration accuracy $e_{\bm{t}} < 0.002173$ and $e_{\bm{R}} < 0.169^{\circ}$ of the \gls{HIL} facility they are set to zero. 
Both error scores combined result in $e_{\text{pose}} = e_{\bm{t}} + e_{\bm{R}}$.
\subsubsection{Challenge Results}
Until the conclusion of the challenge, the ground truth labels of the test set have not been publicly available. 
To add an entry to the challenge, a spreadsheet of the results had to be submitted to the official challenge website. 
A summary of the results of the best submissions is shown in Table~\ref{tab:speedplus_results}. 
%The table is a compilation of each team's \textit{best} result from the official submissions\footnote{\url{https://kelvins.esa.int/pose-estimation-2021/}} to the live and post-mortem challenge or any publication. 
The table is a compilation of each team's \textit{best} result either from the official submission\footnote{\url{https://kelvins.esa.int/pose-estimation-2021/}} tables (live or post-mortem) or any publication.
Our scores are taken from the official post-mortem leaderboard under the team name \textit{mystery\_team}.
Not all top-performing teams have made their method public yet.

Compared to the best submissions on SPEED+, our \method~ achieves overall state-of-the-art. 
In the \texttt{lightbox} category, our \method~ performs best on all three metrics. % $e_{\bm{t}}, e_{\bm{R}}$, and $e_{\text{pose}}$. 
In the \texttt{sunlamp}, our method is third in regards of $e_{\bm{t}}$ and $e_{\text{pose}}$, however performs best in rotational error $e_{\bm{R}}$.
Fig.~\ref{fig:speedplus_samples} shows some qualitative results.
\begin{table}
    \begin{center}
        \caption{Comparison with state-of-the-art on SPEED+. Values depict the best-found results in a publication, the live, or post-mortem challenge.}\label{tab:speedplus_results} 
        \resizebox{\columnwidth}{!}{
        \begin{tabular}[t]{lcccccccc}
            \toprule[1pt]
            & \multicolumn{3}{c}{lightbox}  &\multicolumn{3}{c}{sunlamp} \\ 
            \cmidrule(lr){2-4} \cmidrule(lr){5-7}
             & $e_{\bm{t}}$ & $e_{\bm{R}}$ & $e_{\text{pose}}$  &$e_{\bm{t}}$ & $e_{\bm{R}}$ &$e_{\text{pose}}$ & $\mu$\\
            \cmidrule(lr){1-8}
            \text{lava1302~\cite{Wang2022-rs}} &0.0464	&0.1163	&0.1627 &\textbf{0.0069} &0.0476	&\textbf{0.0545} &0.1086 \\
            \text{prow}                 &0.0196 &0.0944 &0.1140 &0.0133 &0.0840 &0.0972 &0.1056\\
            \text{VPU~\cite{Perez-Villar_undated-ly}}                  &0.0215 &0.0799 &0.1014 &0.0118 &0.0493 &0.0612 &0.0813  \\
            \text{TangoUnchained}       &0.0161 &0.0519 &0.0679 &0.0150  &0.0750 &0.0900 &0.0790\\
            \text{haoranhuang\_njust}   &0.0138 &0.0515 &0.0652 &0.0110 &0.0479 &0.0589 &0.0621\\
            \method~(ours)                     &\textbf{0.0085} &\textbf{0.0305} &\textbf{0.0390} &0.0126 &\textbf{0.0465} &0.0590 &\textbf{0.0490} \\
            \bottomrule[1pt]
        \end{tabular}     
        }
    \end{center}
    \vspace{-4mm}
\end{table}

\subsection{SPEED+ Ablation Studies}\label{exp:speedplus_ablation}
To further investigate some factors of our method and dataset, we present several ablation experiments with a smaller pre-trained ResNet-34~\cite{He2015-hv} as the backbone of \method. 
Moreover, we reduce the size of the decoder layers significantly and use skip connections in each layer. % at each downsampling step of the backbone to the upsampling step of the decoder. 
We trained the model for $1\mathrm{e}6$ steps, using the same optimizer, batch size, and training regiment as in~\ref{subsubsec:exp:training}.
%Moreover, we used the refined object detection bounding boxes from~\ref{subsubsec:exp:object_detection} that we computed for the challenge for all ablation studies.
As input, we consider the refined bounding boxes that we computed for the challenge.
\begin{table}%[b]
    \begin{center}
        \caption{Ablation Study on SPEED+. \method~trained with different targets (normalized model coordinates $\bm{I_{\tilde{q}}}$, object mask $\bm{I}_o$, predicted error $\bm{I}_e$, and surface region segmentation $\bm{I}_r$) and pose estimates with error policies $\varepsilon$.}\label{tab:ablation_error} 
        \begin{tabular}[t]{P{0.01cm} P{0.01cm} P{0.01cm} P{0.01cm} P{0.4cm} P{0.5cm} P{0.5cm} P{0.5cm} P{0.5cm} P{0.5cm} P{0.5cm}}%{ccccc|cccccc}
            \toprule[1pt]
            \multicolumn{5}{c}{} & \multicolumn{3}{c}{lightbox}  &\multicolumn{3}{c}{sunlamp} \\ 
            \cmidrule(lr){6-8} \cmidrule(lr){9-11}
            $\bm{I_{\tilde{q}}}$ & $\bm{I}_o$ & $\bm{I}_e$ & $\bm{I}_r$ & $\epsilon$ & $e_{\bm{t}}$ & $e_{\bm{R}}$ & $e_{\text{pose}}$ &$e_{\bm{t}}$ & $e_{\bm{R}}$ & $e_{\text{pose}}$ \\
            \cmidrule(lr){1-11}
             \cmark & \cmark & \xmark & \xmark & 1.0   &0.037 &0.190 &0.228 &0.056 &0.387 &0.443  \\ [1ex] %baseline
             \cmark & \cmark & \cmark & \xmark & 1.0    &0.043 &0.187 &0.230 &0.070 &0.370 &0.442    \\   %model a epsilon=1.0
             \cmark & \cmark & \cmark & \xmark & $f_{\alpha}$   &0.036 &0.137 &0.174 &0.062 &0.266 &0.328    \\   [1ex]%model a epsilon=$f_{\alpha}$
             \cmark & \cmark & \cmark & \cmark & 1.0    &0.039 &0.161 &0.200 &0.063 &0.327 &0.390    \\ %model b epsilon=1.0
             \cmark & \cmark & \cmark & \cmark & $f_{\alpha}$  &\textbf{0.033} &\textbf{0.117} &\textbf{0.150} &\textbf{0.056} &\textbf{0.237} &\textbf{0.293}    \\ %model b epsilon=$f_{\alpha}$
            \bottomrule[1pt]
        \end{tabular}     
    \end{center}
\vspace{-5mm}
\end{table}

\subsubsection{Effectiveness of Error Prediction and Region Surfaces}\label{subsubsec:exp:error}
We demonstrate the effectiveness of the error prediction map $\widehat{\bm{I}}_e$ and surface region segmentations $\widehat{\bm{I}}_r$. To this end, we train three models: 
\begin{enumerate*}[label=(\roman*)]
    \item a baseline trained with $\bigl[\bm{I_{\tilde{q}}} \conc \bm{I}_o\bigr]$ only,
    \item an error-aware model trained with $\bigl[\bm{I_{\tilde{q}}} \conc \bm{I}_o \conc \bm{I}_e\bigr]$, and
    \item an error-aware, geometry-guided model trained with $\bigl[\bm{I_{\tilde{q}}} \conc \bm{I}_o \conc \bm{I}_e \conc \bm{I}_r\bigr]$.
\end{enumerate*} 
In this experiment, we focus on the quality of \gls{PnP} pose estimates given a set of 2D-3D correspondences and do not use the multi-hypothesis refinement step. 
Moreover, we want to explore the potential of using $\widehat{\bm{I}_e}$ to select different sets of correspondences. 
Hence, we employ an adaptive error threshold $f_{\alpha}(\bm{I}, \widehat{\bm{I}}_e) = \epsilon$ with ${\epsilon \in [1.0, 0.5, 0.3, 0.1, 0.075, 0.05, 0.025]}$ which chooses the best threshold given an image $\bm{I}$ and error map $\widehat{\bm{I}}_e$ statistics. 

As shown in Table~\ref{tab:ablation_error}, the employment of the error prediction output without considering it during \gls{PnP} ($\epsilon = 1.0$) leads to marginally worse metric scores.
However, considering the output to discard erroneous correspondences the performance improves.
%As shown in Table~\ref{tab:ablation_error}, if we add the extra error prediction target $\bm{I}_e$ but do not make any use of it $\epsilon = 1.0$, the performance gets marginally worse. However, if we utilize the error threshold $\epsilon=f_\alpha$, we can improve the accuracy significantly by discarding erroneous correspondences. 
% It should be noted, that $\epsilon=f_\alpha$ should not be considered an upper bound for the performance due to the fixed set of discrete values for $\epsilon$.
Adding the region segmentation as an auxiliary task further improves the results.

\subsubsection{Bridging the Domain Gap with Data Augmentations} \label{exp:speedplus_ablation:domain_gap}
%As mentioned in~\ref{subsubsec:exp:sim2real}, the domain gap was one of the central obstacles to overcome in the dataset. 
As mentioned the Sim2Real gap is one of the central obstacles of SPEED+.
Table~\ref{tab:ablation_augmentations} depicts five models confronted with different data augmentations of increasing severity to investigate their impact on bridging the domain gap.
\begin{table}[b]
    \begin{center}
        \vspace{-3mm}
        \caption{Ablation Study on Different Data Augmentations Configurations on SPEED+.}\label{tab:ablation_augmentations} 
        \begin{tabular}[t]{lcccccc}
            \toprule[1pt]
            & \multicolumn{3}{c}{lightbox}  &\multicolumn{3}{c}{sunlamp} \\ 
            \cmidrule(lr){2-4} \cmidrule(lr){5-7}
             & $e_{\bm{t}}$ & $e_{\bm{R}}$ & $e_{\text{pose}}$ &$e_{\bm{t}}$ & $e_{\bm{R}}$ & $e_{\text{pose}}$ \\
            \cmidrule(lr){1-7}
             \texttt{aug1}   &0.050 &0.216 &0.266 &0.122 &0.557 &0.679  \\
             \texttt{aug2}   &0.033 &0.121  &0.154  &0.058 &0.223  &0.281   \\
             \texttt{aug3}   &0.039 &0.138  &0.178  &0.034  &0.142  &0.176    \\
             \texttt{aug4}   &0.038 &0.128  &0.166  &\textbf{0.033}  &\textbf{0.133}  &\textbf{0.166}    \\
             \texttt{aug5}   &\textbf{0.032} &\textbf{0.112}  &\textbf{0.145}  &0.035  &0.138  &0.173   \\
            \bottomrule[1pt]
        \end{tabular}     
    \end{center}
    \vspace{-4mm}
\end{table}
%To investigate the impact of different data augmentation modules, we trained five models with a different set of data augmentations of increasing severity. 
Therefore, we train
\begin{enumerate*}[label=(\roman*)]
    \item \texttt{aug1} with blur, sharpen, emboss, additive gaussian noise, coarse dropout, linear contrast, add, invert, and multiply,
    \item \texttt{aug2} increases the number of augmentations and adds superpixels, edge detections, and dropout,
    \item \texttt{aug3} adds simplex noise, image corruptions (fog, snow), and space-specific augmentations (specular reflection, high contrast),
    \item \texttt{aug4} adds an extra brightness module and further image corruptions (saturation, contrast),
    \item at last \texttt{aug5} separates the different classes of augmentations into brightness, blur, corruptions, and general augmentations and samples these groups individually.\footnote{A full list of data augmentations can be found in our github repository (soon available)}
\end{enumerate*} 

One can observe the vital role of data augmentation in addressing the Sim2Real gap~\cite{Sundermeyer2018-fl}. 
% As shown in Table~\ref{tab:ablation_augmentations}, data augmentations play a vital role in bridging the domain gap. 
The results also indicate that test data-inspired augmentations, such as synthetic specular reflections, can drastically improve performance. % of a model. 
This can be especially seen in the \texttt{sunlamp} test set. 
At last, we see a difference between \texttt{aug4} and \texttt{aug5} which mainly differ in additional logic applying augmentations. 
% We hypothesize that adding too many augmentations of the same type can deteriorate the information contained in a sample to the point that no information remains. 
Model capacity plays an integral role in the selection and amount of data augmentations.
Hence, while this study might offer some clues, future research is required to find definitive factors in data augmentations for orbital perception.
\begin{figure}
    \centering
    \includegraphics[width=0.45\textwidth]{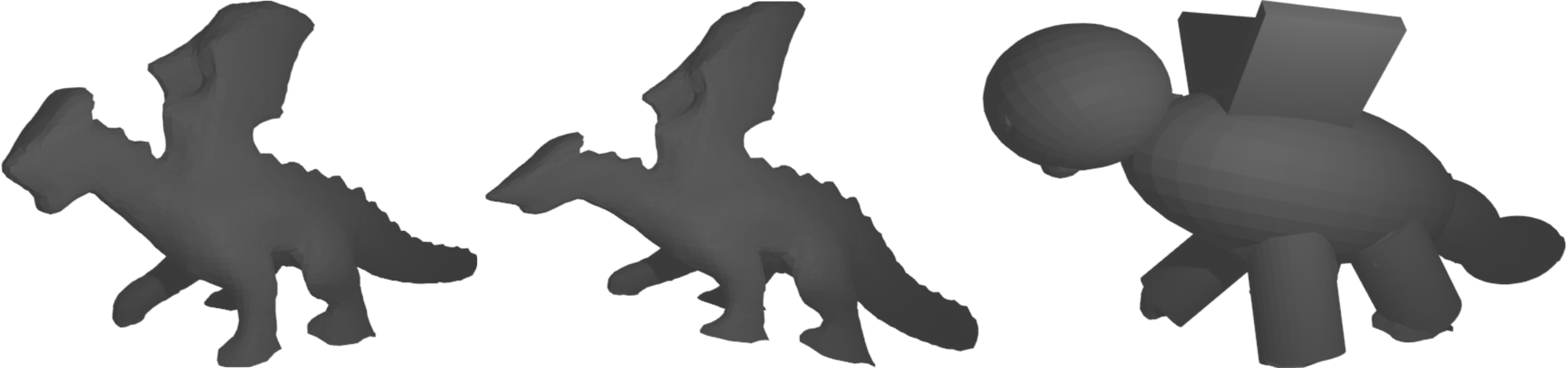}
    \caption{Illustration of the three models $q_1, q_2$, and $q_3$ that we use in the model quality ablation study. (left) ground truth 3D CAD model; (middle) deformed model; (right) model of geometric primitives.}
    \label{fig:tudl_dragons}
    \vspace{-5mm}
\end{figure}

\subsubsection{Multi-hypothesis Testing and Refinement}
Finally, we demonstrate the effects of refinement, multi-hypotheses testing, and test set augmentations. 
For this study, the \texttt{aug5} model is used and %trained for the data augmentation ablation study is used. 
%We investigate several different scenarios:
several scenarios are investigated:
\begin{enumerate*}[label=(\roman*)]
    \item no refinement with $\epsilon=1.0$,
    \item single hypothesis refinement with $\epsilon=1.0$,
    \item multi-hypothesis refinement of all hypotheses with predicted, 
    \item ground truth, or %multi-hypothesis refinement of all available hypothesis with the ground truth segmentation $\bm{I}_o$, 
    \item no learned object segmentation map; %multi-hypothesis refinement without the learned confidence map,
    \item multi-hypothesis refinement with predicted segmentation and test set augmentations.
\end{enumerate*}

\begin{table}[b]
\vspace{-4mm}
    \begin{center}
        \caption{Ablation Study on Different Refinement Design Choices on SPEED+.}\label{tab:ablation_refinement} 
        \vspace{-2mm}

        \begin{tabular}[t]{lcccccc}
            \toprule[1pt]
            & \multicolumn{3}{c}{lightbox}  &\multicolumn{3}{c}{sunlamp} \\ 
            \cmidrule(lr){2-4} \cmidrule(lr){5-7}
              & $e_{\bm{t}}$ & $e_{\bm{R}}$ & $e_{\text{pose}}$ &$e_{\bm{t}}$ & $e_{\bm{R}}$ & $e_{\text{pose}}$ \\
            \cmidrule(lr){1-7}
             (i) no refinement   &0.038   &0.157  &0.194  &0.038  &0.192  &0.230  \\
             (ii) single $h_i$   &0.035 &0.133 &0.168 &0.036 &0.164 &0.200   \\
             (iii) multi $h_i$; $\widehat{\bm{I}}_o$   &0.030	&0.126	&0.156	&0.032 &0.159 &0.191    \\
             (iv) multi $h_i$; $\bm{I}_o$    &0.024 &0.104 &0.127 &0.024 &0.138 &0.162   \\
             (v) without $\bm{I}_o$    &0.037 &0.166 &0.203 &0.042 &0.257 &0.299 \\
             (vi) test set augs     &\textbf{0.019}	&\textbf{0.088}	&\textbf{0.107} &\textbf{0.022} &\textbf{0.097} &\textbf{0.120}    \\
            \bottomrule[1pt]
        \end{tabular}     
    \end{center}
\vspace{-4mm}
\end{table}

As shown in Table~\ref{tab:ablation_refinement}, the refinement significantly improves the prediction accuracy, even with only one hypothesis ((i) vs. (ii)). 
This improvement increases even further if multiple-hypothesis refinement is applied.
However, the refinement performance directly correlates with the quality of the segmentation map ((iii) vs. (iv)).
At last, the (vi) test set augmentations drastically improve the accuracy by another $35\%$ on average. 
Besides the fact, that we refine four times more hypotheses, the ensemble significantly improves the quality of the segmentation mask $\widehat{\bm{I}}_o = \frac{1}{4}\sum_{i=1}^4 \widehat{\bm{I}}_{o,i}$. 

\subsection{Experiments on TUD-L Dataset}
In the last experiment, we demonstrate the impact of object model quality on correspondence-based pose estimation methods. 
More importantly, we showcase how \method~ is still able to make accurate predictions.
For this experiment, we use the TUD-L dataset~\cite{hodan_bop_2018} -- similar to SPEED+ -- contains challenging lighting effects on the real test set. 
\begin{figure}
    \centering
    \includegraphics[width=0.48\textwidth]{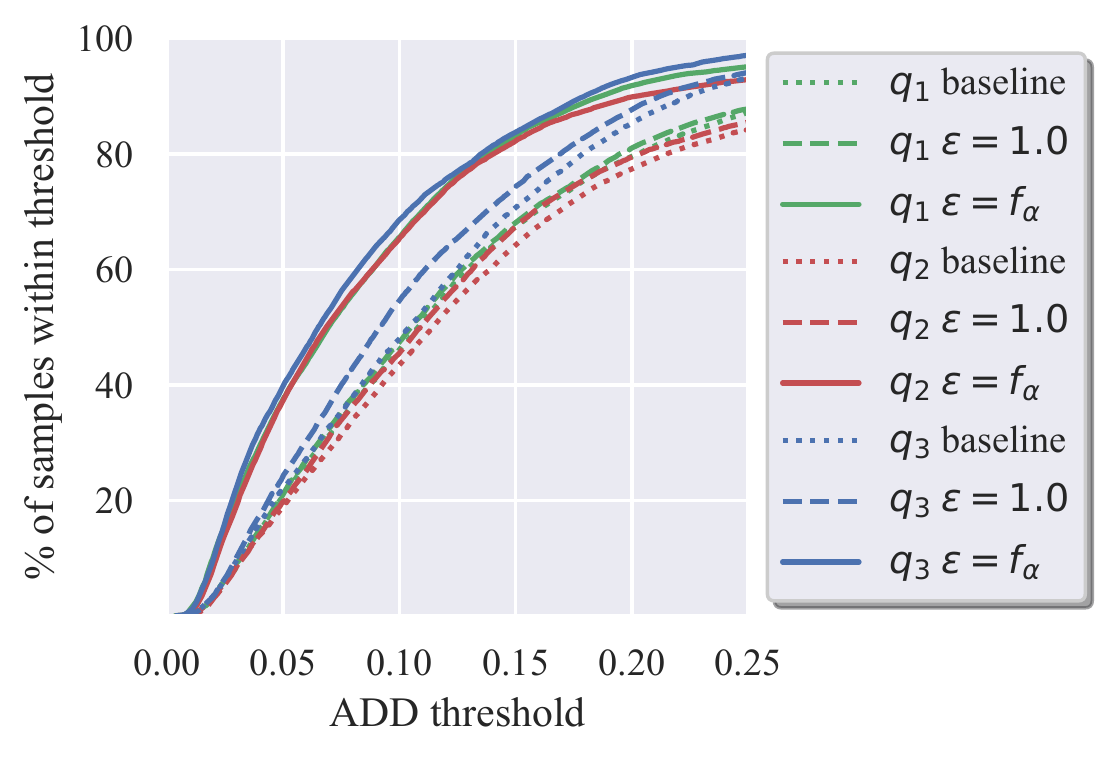}
    \caption{Results of the ablation study on TUD-L. The graph shows the fraction of samples within ADD thresholds. The larger the area under the curve, the better the performance. Methods using $\varepsilon=f_\alpha$ clearly outperform the other methods and can accurately estimate the pose.}\label{fig:tudl_ablation}
\vspace{-4mm}
\end{figure}

To investigate the impact of the object's model quality, we create two approximate object models (see Fig.~\ref{fig:tudl_dragons}). 
The first model $q_1$ acts as a baseline and is not changed. 
For the second model $q_2$, we taper the model using Blenders \texttt{Taper} modifier. 
Such a model deterioration could arise if the 3D model used for learning is generated using 3D shape reconstruction. 
At last, the model $q_3$ is approximating the real object with coarse geometric primitives which corresponds to a reduced modelling effort. 
To quantify the model deformity, we calculate distances from vertices in $q_1$ to the nearest neighbor vertices in $q_2$ and $q_3$. On average, $q_2$ has a distance error of $5 \mathrm{mm}$ per model point and model $q_3$ of $7 \mathrm{mm}$.

We train the models purely on synthetic images generated with BlenderProc~\cite{denninger_blenderproc_nodate} with the same training parameters as in \ref{exp:speedplus_ablation} and basic data augmentations \texttt{aug2}. For each object quality we train two models,
\begin{enumerate*}[label=(\roman*)]
    \item a \textbf{baseline} with $\bigl[\bm{I_{\tilde{q}}} \conc \bm{I}_o\bigl]$, and
    \item an \textbf{error-aware} model with $\bigl[\bm{I_{\tilde{q}}} \conc \bm{I}_o \conc \bm{I}_e\bigl]$.
\end{enumerate*} 

We use the ADD metric for evaluation given the ground truth model. It measures whether the average distances of the transformed model points deviate less than a certain threshold of the object's diameter. As shown in Fig.~\ref{fig:tudl_ablation}, error-aware models that use the adaptive error threshold $\epsilon=f_\alpha$ perform significantly better than the other methods. They achieve an ADD$(0.1\mathrm{d})$, less than $10$\% of their diameter, for between $65$\% and $70$\% of the test images. In contrast, the baseline models achieved an ADD$(0.1\mathrm{d})$ between $43$\% and $46$\%. Surprisingly, error-aware models with $\epsilon=1.0$ significantly outperform the baseline methods with an ADD$(0.1\mathrm{d})$ between $45$\% and $55$\% without using the estimated errors.

%% file: sections/conclusions.tex
\section{Conclusion}
In this work, we presented \method, a dense 2D-3D correspondence model, and multi-hypothesis refinement framework to estimate accurate 6D poses under very challenging visual conditions, such as aggressive reflections and low signal-to-noise ratio. 
We have shown the benefit of estimating coordinate prediction errors which we used to formulate multiple pose hypotheses using \gls{PnP}. 
Each individual pose hypothesis is refined and the maximum probability estimate yields the final pose. 
Our approach achieves state-of-the-art performance on the challenging SPEED+ dataset and post-mortem SPEC2021 competition. 
Finally, we demonstrated the benefit of coordinate prediction errors for pose estimation with approximated 3D models on the TUD-L dataset.
% The main limitation of this work is \gls{PnP} which currently generates a huge computational overhead. Due to this, the system is currently not real-time capable. Moreover,   
